\PassOptionsToPackage{usenames,dvipsnames,table,hyperref}{xcolor}
\documentclass[11pt,a4paper]{article}
\usepackage[hyperref]{naaclhlt2018}
\usepackage{latexsym}
\usepackage[utf8]{inputenc} % allow utf-8 input
\usepackage[T1]{fontenc}    % use 8-bit T1 fonts
\usepackage{hyperref}       % hyperlinks
\usepackage{url}            % simple URL typesetting
\usepackage{booktabs}       % professional-quality tables
\usepackage{amsfonts}       % blackboard math symbols
\usepackage{nicefrac}       % compact symbols for 1/2, etc.
\usepackage{microtype}      % microtypography
\usepackage{latexsym}
\usepackage{amssymb}
\usepackage{times}
\usepackage{latexsym}
\usepackage{amsmath}
\usepackage{multirow}
\usepackage{amsfonts}
\usepackage{booktabs}
\usepackage{rotating}
\usepackage{subfig}
\usepackage{array}
\usepackage{booktabs}
\usepackage{amsfonts}
\usepackage{enumitem}% http://ctan.org/pkg/enumitem
\usepackage{makecell}
\usepackage{xcolor}
\usepackage[toc,page]{appendix}
\usepackage{amsmath}
\usepackage{xspace}
\usepackage{mdwlist}
\usepackage{graphics}
\usepackage{color}
\usepackage{rotating}
\usepackage{booktabs}
\usepackage{epsfig}
\usepackage{alltt}
\usepackage{moreverb}
\usepackage{fancyvrb}
\usepackage{enumerate}
\usepackage{colortbl}      % for color table cells
\usepackage{xspace}
\usepackage{relsize}
\usepackage{array}
\usepackage{amsmath}
\usepackage{amsfonts}
\usepackage{txfonts}
\usepackage{caption}
\DeclareCaptionFont{10pt}{\fontsize{10pt}{12pt}\selectfont}
\usepackage{rotating}
\usepackage{latexsym}
\usepackage{multirow}
\usepackage{booktabs}
\usepackage{float}
\usepackage{array,etoolbox}
\usepackage{enumitem}
\usepackage{subfig}
\usepackage{arydshln}
\usepackage{setspace}
\usepackage{float}
\usepackage{mathtools}
\usepackage{bm}
\usepackage{xcolor}
\usepackage[toc,page]{appendix}
\usepackage{soul}
\usepackage[draft]{pgf}
\usepackage{makecell}
\usepackage[normalem]{ulem}
\usepackage{amssymb}
\usepackage{wasysym,textcomp}
\usepackage[toc,page]{appendix}
\usepackage{wrapfig}

\usepackage{tabularx}
\definecolor{vlgray}{gray}{0.95}

\usepackage{etoolbox}
\newtoggle{draft}
\toggletrue{draft}
\iftoggle{draft}{
\newcommand{\peter}[1]{\textcolor{Blue}{[#1 \textsc{--Pete}]}}
\newcommand{\scott}[1]{\textcolor{Maroon}{[#1 \textsc{--Scott}]}}
\newcommand{\lifu}[1]{\textcolor{Orange}{[#1 \textsc{--Lifu}]}}
\newcommand{\bhavana}[1]{\textcolor{Green}{[#1 \textsc{--Bhavana}]}}
}{
\newcommand{\peter}[1]{}
\newcommand{\scott}[1]{}
\newcommand{\lifu}[1]{}
\newcommand{\bhavana}[1]{}
}

 \newcommand{\finaledit}[1]{#1}

\newcommand{\com}[1]{}
\newcommand{\myparagraph}[1]{\vspace{1mm} \noindent {\bf #1: }}
\newcommand{\bfit}[1]{\textbf{\textit{#1}}}
\newcommand{\eat}[1]{}
\mathchardef\mhyphen="2D
\newenvironment{ite}{                     % list without par spacings
     \parskip 0cm \begin{itemize} \parskip 0cm \parsep 0cm \itemsep 0cm \topsep 0cm}{
        \end{itemize}} %  \parskip 0cm}
\newenvironment{enu}{                   % list without par spacings
     \parskip 0cm \begin{list}{}{\parsep 0cm \itemsep 0cm \topsep 0cm}}{
       \end{list}} %  \parskip 0cm}
\newenvironment{des}{                 % list without par spacings
     \parskip 0cm \begin{list}{}{\parsep 0cm \itemsep 0cm \topsep 0cm}}{
       \end{list}} %  \parskip 0cm}
\newcommand{\prolocal}{\textsc{ProLocal}}
\newcommand{\proglobal}{\textsc{ProGlobal}}

\makeatletter
\g@addto@macro\normalsize{%
  \setlength\abovedisplayskip{1pt}
  \setlength\belowdisplayskip{1pt}
  \setlength\abovedisplayshortskip{1pt}
  \setlength\belowdisplayshortskip{1pt}
}
\makeatother

\newcommand{\dataset}{ProPara}

\aclfinalcopy

\title{Tracking State Changes in Procedural Text: \\A Challenge Dataset and Models for Process Paragraph Comprehension}

\author{
\makecell{Bhavana Dalvi Mishra$^1$\textsuperscript{*},\  Lifu Huang$^2$\thanks{\textsuperscript{*}Bhavana Dalvi Mishra and Lifu Huang contributed equally to this work.},\ \  Niket Tandon$^1$, Wen-tau Yih$^1$, Peter Clark$^1$} \\
$^1$Allen Institute for AI, Seattle, $^2$Rensselaer Polytechnic Institute, Troy\\
{\tt $\{$bhavanad,nikett,scottyih,peterc$\}$@allenai.org, $\{$warrior.fu$\}$@gmail.com}\\
}

\date{}

\hypersetup{draft}
\begin{document}
\maketitle

\begin{abstract}
We present a new dataset and models for comprehending paragraphs about 
processes (e.g., photosynthesis), an important genre of text describing a dynamic world.
The new dataset, ProPara, is the first to  contain natural (rather than machine-generated) text about a changing world 
along with a full annotation of entity states (location and existence) during those changes (81k datapoints).
The end-task, tracking the location and existence of entities through the text,
is challenging because the causal effects of actions are often
implicit and need to be inferred. We find that previous models that
have worked well on synthetic data achieve only mediocre 
performance on ProPara, and introduce two new neural models that exploit 
alternative mechanisms for state prediction, in particular
using LSTM input encoding and span prediction. The new models
improve accuracy by up to 19\%. 
\finaledit{The dataset and models are available to the community at \small \url{http://data.allenai.org/propara}}.
% We are releasing the ProPara dataset and our models\footnote{The ProPara dataset can be downloaded from \url{http://data.allenai.org/propara}. Our model code is available at \url{https://github.com/allenai/propara}. } to the community. 

\end{abstract}

\section{Introduction} 

\eat{Building a reading comprehension (RC) system that is able to read a text document and to answer 
questions accordingly has been a long-standing goal in NLP and AI research.  Despite the fact that 
impressive progress has been made, enabled by well-designed datasets and modern neural network 
models, state-of-the-art RC systems thrive on answering simple questions by sophisticated pattern 
recognition, but still struggle with questions requiring \emph{inference}~\cite{jia2017adversarial}. }

Building a reading comprehension (RC) system that is able to read a text document and to answer 
questions accordingly has been a long-standing goal in NLP and AI research. Impressive progress has
been made in factoid-style reading comprehension, e.g.,~\cite{Seo2016BidirectionalAF,clark2017simple}, enabled by well-designed datasets and modern neural
network models. However, these models still struggle with questions that require \emph{inference}~\cite{jia2017adversarial}.

Consider the paragraph in Figure~\ref{fig:example} about photosynthesis. While top systems on 
SQuAD~\cite{rajpurkar2016squad} can reliably answer lookup questions such as:

\noindent 
\textbf{Q1}: What do the roots absorb? (A: water, minerals)

\noindent
they struggle when answers are not explicit, e.g.,

\noindent
\textbf{Q2}: Where is sugar produced? (A: in the leaf)\footnote{For example, the RC system BiDAF~\cite{Seo2016BidirectionalAF} answers ``glucose'' to this question.}

\noindent
To answer Q2, it appears that a system needs knowledge of the world and the ability to reason 
with state transitions in multiple sentences: 
If carbon dioxide {\it enters} the leaf (stated), then it will be {\it at}
the leaf (unstated), and as it is then used to produce sugar, the sugar
production will be at the leaf too. 

\begin{figure}
\centerline{
\fbox{%
    \parbox{0.44\textwidth}{%
	Chloroplasts in the leaf of the plant trap light from the sun.
	The roots absorb water and minerals from the soil.
	This combination of water and minerals flows from the stem into the 
	leaf. Carbon dioxide enters the {\bf leaf}.
	Light, water and minerals, and the carbon dioxide all combine into a mixture.
	This mixture forms {\bf sugar} (glucose) which is what the plant eats.
% {\bf Q:} Where is sugar produced?(A:the leaf)
\vspace{2mm}
\begin{des}
\item[{\bf Q:}] Where is sugar produced?
\item[{\bf A:}] in the leaf
\end{des}
    }%
}}
\caption{A paragraph from \emph{ProPara} about photosynthesis (bold added, to
highlight question and answer elements). Processes are challenging because 
questions (e.g., the one shown here) often require inference about the process states.}
% often require inference (implicit answers)
% A challenging question (as the answer is implicit) on a paragraph about photosynthesis.}
% Inference is required to answer this question from a paragraph about photosynthesis.}
\label{fig:example}
\vspace{-5mm}
\end{figure}

%--------------------------------------------------------------------

This challenge of modeling and reasoning with a changing world is particularly pertinent in 
text about {\it processes}, demonstrated by the paragraph in Figure~\ref{fig:example}. 
% 
% SY: Probably need to revise or remove the following sentences.
%     But the goal here is to emphasize the generality of the ``processes'' problem.
Understanding what is happening in such texts is important for many tasks, e.g.,
procedure execution and validation, effect prediction.
However, it is also difficult because the world state is changing, and
the causal effects of actions on that state are often implicit.

%--------------------------------------------------------------------

% SY: The text below should justify why we need another dataset.
%% 1. Issues of the existing datasets
%% 2. The pros of our dataset

%% SY: Could we have a name for the reasoning problem we are addressing here?
To address this challenging style of reading comprehension problem, 
% requiring reasoning about a dynamic world, 
researchers have created several datasets.
The bAbI dataset~\cite{weston2015towards} includes questions about 
objects moved throughout a paragraph, using machine-generated language over
a deterministic domain with a small lexicon. The SCoNE dataset~\cite{long2016simpler} 
contains paragraphs describing a changing world state in three synthetic, 
deterministic domains, and assumes that a complete and correct model of the initial 
state is given for each task.  However, approaches developed using synthetic
data often fail to handle the inherent complexity in language when applied
to organic, real-world data~\cite{hermann2015teaching,winograd72}.

%% SY: We need to say something about the dataset from [Berant et al., 2014]

\begin{figure}
\centering
\includegraphics[width=\columnwidth]{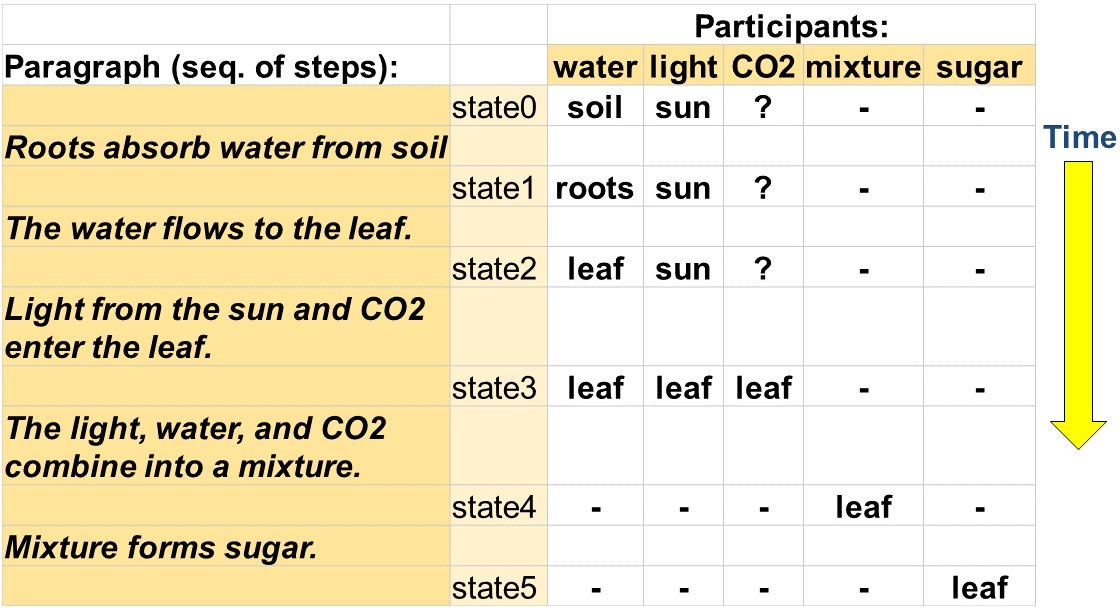}
\caption{A (simplified) annotated paragraph from ProPara. Each filled row shows the 
existence and location of participants between each step (``?'' denotes ``unknown'', ``-'' denotes ``does not exist''). For example in state0, water is located at the soil.}
\vspace{-0.4mm}
\label{participant-grid}
\end{figure}

In this work, we create a new dataset, \emph{ProPara} (Process Paragraphs), containing 
488 human-authored paragraphs of procedural text, along with 
81k annotations about the changing states (existence and location) of entities in those paragraphs,
with an end-task of predicting location and existence changes that occur.
This is the first dataset containing annotated, natural text for 
real-world processes, along with a simple representation of entity states 
during those processes. A simplified example is shown in Figure~\ref{participant-grid}.

%% SY: What are the unique advantages of ProPara?

%--------------------------------------------------------------------

When applying existing state-of-the-art systems, such as Recurrent Entity Networks~\cite{Henaff2016TrackingTW} 
and Query-reduction Networks~\cite{Seo2017QueryReductionNF}, we find that they do not perform well
on \emph{ProPara} and the results are only slightly better than the majority baselines. 
As a step forward, we propose two new neural models that use alternative mechanisms for state prediction and propagation,
in particular using LSTM input encoding and span prediction. The new models improve accuracy by up to 19\%.
\eat{
As a step forward, we propose two new neural models that use alternative mechanisms for state prediction and propagation. 
In particular, for state values expected to be mentioned in the paragraph,
we use a span prediction approach (found useful by many QA systems for SQuAD-style data), resulting in accuracy improvements of up to 19\%.}
% rather than classification approach, which results in improvement of accuracy by up to 19\%. 

%--------------------------------------------------------------------

Our contributions in this work are twofold: (1) we create \emph{ProPara}, a new dataset for 
process paragraph comprehension, containing annotated, natural language paragraphs about 
real-world processes, and (2) we propose two new models that learn to infer and propagate 
entity states in novel ways, and outperform existing methods on this dataset.

%--------------------------------------------------------------------

%% SY: I usually add the structure paragraph in the end of intro.
%%	   The rest of this paper is structured as follows...

%--------------------------------------------------------------------

\section{Related Work \label{related-work}}

% We survey related datasets and existing models proposed for comprehending procedural text below.

%\subsection{Datasets}
% \noindent \paragraph{Datasets:}
{\bf Datasets:}
Large-scale reading comprehension datasets, e.g., SQuAD \cite{rajpurkar2016squad}, 
TriviaQA \cite{JoshiTriviaQA2017}, have successfully driven progress in question answering,
but largely targeting explicitly stated facts.  Often, the resulting systems
can be fooled~\cite{jia2017adversarial}, prompting
efforts to create harder datasets where a deeper understanding 
of the text appears necessary~\cite{welbl2017constructing,arakigenerating}.

Procedural text is a genre that is particularly challenging,
because the worlds they describe are largely implicit and changing.
While there are few large datasets in this genre, two exceptions are 
bAbI~\cite{weston2015towards} and SCoNE~\cite{long2016simpler},
described earlier\footnote{
The ProcessBank \cite{berant2014modeling} dataset is smaller and does 
not address state change, instead containing 585 questions about 
event ordering and event arguments.}.  bAbI has helped advance methods for reasoning 
over text, such as memory network architectures~\cite{weston2014memory},
but has also been criticized for using machine-generated text over a
simulated domain. SCoNE is closer to our goal, but has a different
task ({\it given} a perfect world model of the initial state, predict 
the end state) and different motivation (handling ellipsis and
coreference in context). It also used a deterministic,
simulated world to generate data.

%% SY: Mention the dataset in ProRead? Good idea! Done

%\subsection{Models \label{related-work:models}}
% \noindent \paragraph{Models:}
\noindent {\bf Models:}
For answering questions about procedural text, early
systems attempted to extract a process structure (events, arguments, 
relations) from the paragraph, 
e.g., ProRead \cite{berant2014modeling} and for newswire
\cite{caselli2017proceedings}. This allowed questions about event 
ordering to be answered, but not about state changes, unmodelled by
these approaches.

%% SY: How should we describe Antoine Bosselut's work?
%%     https://arxiv.org/abs/1711.05313
More recently, several neural systems have been developed to
answer questions about the world state after a process, inspired in part
by the bAbI dataset. Building on the general Memory Network architecture \cite{weston2014memory}
and gated recurrent models such as GRU~\cite{cho2014properties},
Recurrent Entity Networks (EntNet)~\cite{Henaff2016TrackingTW} is a
state-of-the-art method for bAbI.  EntNet uses a dynamic memory of hidden 
states (memory blocks) to maintain a representation of the world state, with a gated update
at each step. Memory keys can be preset ("tied") to particular entities in the text, to
encourage the memories to record information about those entities.
% Memories undergo a gated update after each sentence is read, allowing
% states to be propogated (when appropriate) through time.
Similarly, Query Reduction Networks (QRN)~\cite{Seo2017QueryReductionNF} 
tracks state in a paragraph, represented as a hidden vector $h$. QRN performs
gated propagation of $h$ across each time-step (corresponding to a state update), 
and uses $h$ to modify (``reduce'') the query to keep pointing to the answer
at each step (e.g., ``Where is the apple?'' at step$_{1}$ might be modified
to ``Where is Joe?'' at step$_{2}$ if Joe picks up the apple).
\finaledit{A recent proposal, Neural Process Networks (NPN)~\cite{bosselut2017simulating}, also models
each entity's state as a vector (analogous to EntNet's tied memories).
NPN computes the state change at each step from the step's predicted action and affected
entity(s), then updates the entity(s) vectors accordingly, but does not
model different effects on different entities by the same action.}

Both EntNet and QRN find a final answer by decoding the final vector(s) into a 
vocabulary entry via softmax classification. In contrast, many of the best 
performing factoid QA systems, e.g., \cite{Seo2016BidirectionalAF,clark2017simple},
return an answer by finding a {\it span} of the original paragraph using attention-based 
span prediction, a method suitable when there is a large vocabulary. We combine this span prediction
approach with state propagation in our new models.

\section{The ProPara Dataset}
\label{dataset}

\myparagraph{Task}
Our dataset, \emph{ProPara}, focuses on a particular genre of procedural text, namely
simple scientific processes (e.g., photosynthesis, erosion).
A system that understands a process paragraph should be able
to answer questions such as: ``\textit{What are the inputs to the process?}'', 
``\textit{What is converted into what?}'', and ``\textit{Where does the conversion take place?}''\footnote{For example, science exams 
pose such questions to test student's understanding of the text in various ways.}
Many of these questions reduce to understanding the
basic dynamics of entities in the process, and we use this as our task: 
Given a process paragraph and an entity $e$ mentioned in it, identify:
\noindent
(1) {\bf Is} $e$ created (destroyed, moved) in the process?

\noindent
(2) {\bf When} (step \#) is $e$ created (destroyed, moved)?

\noindent
(3) {\bf Where} is $e$ created (destroyed, moved from/to)?
\noindent
If we can track the entities' {\it states} through the process and
answer such questions, many of the higher-level questions can be
answered too. To do this, we now describe how these states
are representated in ProPara, and how the dataset was built.

\myparagraph{Process State Representation}
The states of the world throughout the whole process are represented as a grid.  Each column denotes 
a \emph{participant} entity (a span in the paragraph, typically a noun phrase) that undergoes some 
creation, destruction, or movement in the process.  Each row denotes the \emph{states} of all the 
participants after a \emph{step}.  Each sentence is a step that may change the state of one or more participants.
Therefore, a process paragraph with $m$ sentences and $n$ participants will result in an $(m+1) \times n$ 
grid representation. Each cell $l_{ij}$ in this grid records the \emph{location} of the $j$-th participant
after the $i$-th step, and $l_{0j}$ stores the location of $j$-th participant before the process.\footnote{We only trace 
locations in this work, but the representation can be easily extended to store other properties (e.g., temperature) 
of the participants.}  Figure~\ref{participant-grid} shows one example of this representation.

\myparagraph{Paragraph Authoring}
To collect paragraphs, we first generated a list of 200 process-evoking prompts,
such as ``\emph{What happens during photosynthesis?}'', by instantiating five patterns\footnote{
    The five patterns are: How are \bfit{structure} formed? How does \bfit{system} work? 
    How does \bfit{phenomenon} occur? How do you use \bfit{device}? What happens during \bfit{process}?
}, with nouns of the corresponding type from a science vocabulary, followed by
manual rewording.
Then, crowdsourcing (MTurk) workers were shown one of the prompts and 
asked to write a sequence of event sentences describing the process.
Each prompt was given to five annotators to produce five (independent) paragraphs.
Short paragraphs (4 or less sentences) were then removed for a final total of 
488 paragraphs describing 183 processes. An example paragraph is the one shown earlier in 
Figure~\ref{fig:example}.

\myparagraph{Grid and Existence}
Once the process paragraphs were authored, we asked expert annotators\footnote{
Expert annotators were from our organization, with a college or higher degree.} 
to create the initial grids.  First, for each paragraph, they listed the participant 
entities that underwent a state change during the process, thus 
creating the column headers. 
They then marked the steps where a participant was created or destroyed.
All state cells before a Create or after a Destroy marker were labeled as "not exists". 
Each initial grid annotation was checked by a second expert annotator.

\eat{
They then marked the steps where an participant was
created (C) or destroyed (D), as shown in Figure~\ref{participant-grid-partially-filled}.
From these markings, all state cells before a {\bf C}reate or after a {\bf D}estroy 
were labeled as "not exists". Each initial grid annotation is checked by a second
expert annotator.

\begin{figure}
\centering
\includegraphics[width=\columnwidth]{./figures/participant-grid-partially-filled.JPG}
\caption{In step 1, annotators labeled steps where an entity is
destroyed/consumed ({\bf D}) or created ({\bf C}). From this, all
the ``not exists'' state annotations (all the ``-'' in Figure~\ref{participant-grid})
can be filled in, i.e., all states after {\bf D} or before {\bf C}.}
\label{participant-grid-partially-filled}
\end{figure}
}

\begin{table}
\centering
\scalebox{0.89}{%
\begin{tabular}{|l|l|l|l|} \hline
 & {\bf bAbI} & {\bf SCoNE} & {\bf ProPara} \\ \hline
Sentences & Synthetic & Natural & Natural \\
Questions & templated & templated & templated \\
\# domains & 20 & 3 & 183 \\
Vocab \#words & 119 & 1314 & 2501 \\
% \# qn types & 20 & 3 & 10 \\
% \# qns (total) & 51.7k & 609.1k & 84.1k \\
% \# qns (unique) & 439 & 609.1k & 16.6k \\
% Av \# qns/para & 4.4 & 41.8 & 172.3 \\ 
\# sentences & 131.1k & 72.9k & 3.3k \\
\# unique sents & 3.2k & 37.4k & 3.2k \\
Avg words/sent & 6.5 & 10.2 & 9.0 \\ \hline
% Av. words/para & 69.6 & 39.4 & 53.4 \\
% Av. sents/para & 11.2 & 5.0 & 6.8 \\ \hline
\end{tabular}
}
\caption{ProPara vs. other procedural datasets. \label{dataset-comparison}}
\vspace{-5mm}
\end{table}

\begin{figure*}[tbh]
\centering
\includegraphics[width=1.07\textwidth]{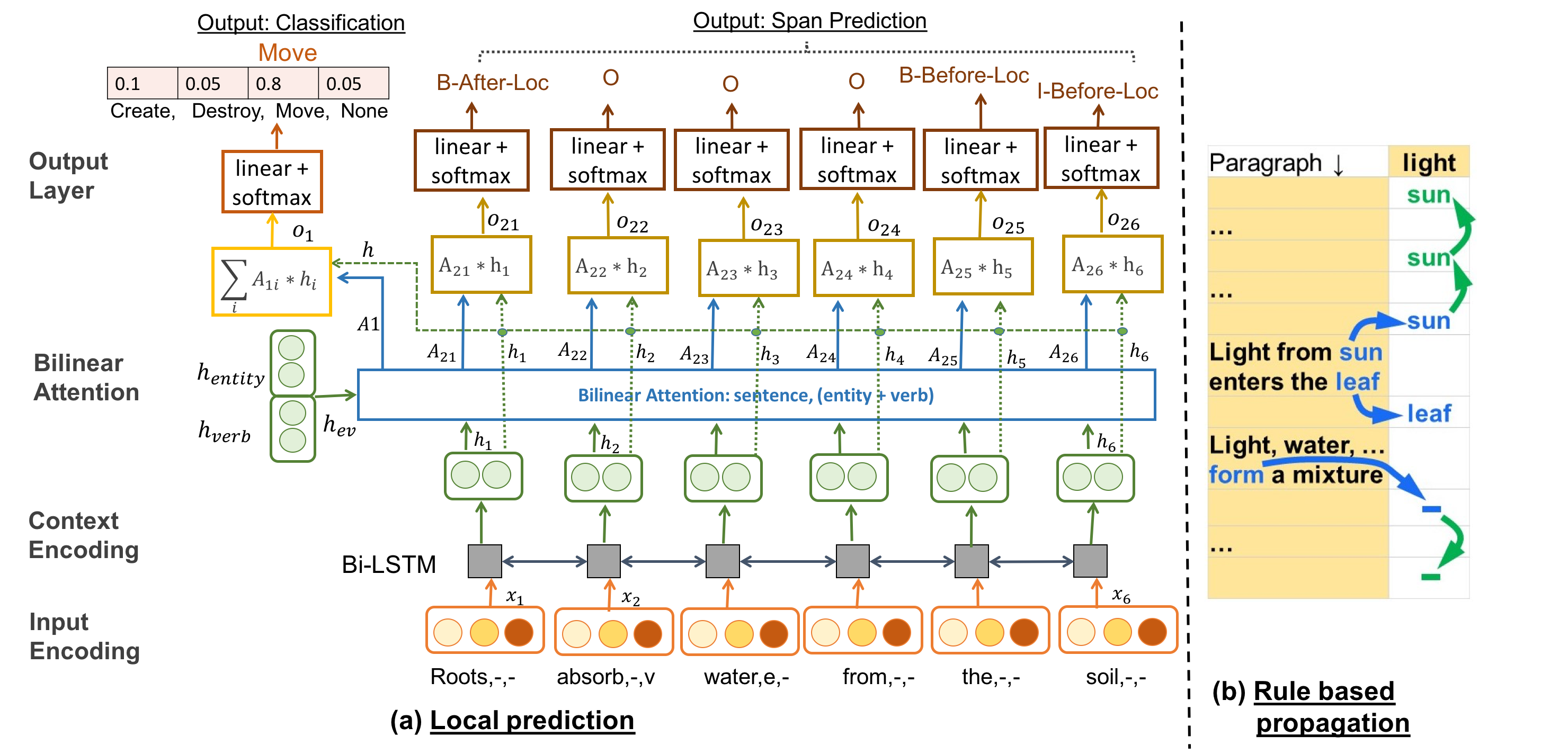}
% \caption{(left) Neural Architecture for \prolocal. (right) Illustration of rule based propagation.  }
\vspace{-5mm}
\caption{(a) \prolocal~uses bidirectional attention to make local predictions about state change type and location (left), and then \finaledit{(b)  propagates those changes globally using a persistence rule (right, shown for a single participant (the Light), local predictions shown in blue, propagations via persistence in green).}}
\label{model1}
\vspace{-1mm}
\end{figure*}

\myparagraph{Locations}
Finally, MTurk workers were asked to fill in all the location cells. 
A location can be ``unknown" if it is not specified in the text, or  
a span of the original paragraph. Five grids for the same paragraph were completed by
five different Turkers, with average pairwise inter-annotator agreement of $0.67$.
The end result was 81,345 annotations over 488 paragraphs about 183 processes.
The dataset was then split 80/10/10 into train/dev/test by {\it process prompt},
ensuring that the test paragraphs were all about processes unseen in train and dev.
% \textbf{Specifically, 25\% of the test vocabulary was unseen in the train/dev partitions. Also, there are 519 different verbs (999 verb phrases) in ProPara. 23\% of the verbs (40\% verb phrases) in the test partition are not present in train or dev, making the task challenging and requiring cross-verb generalization.}
Table~\ref{dataset-comparison} compares our dataset with bAbI and SCoNE.

% A comparison of ProPara to bAbI and SCoNE is shown in Table~\ref{dataset-comparison}.

\vspace{-1mm}

\section{Models}

We present two new models for this task. The first, \prolocal, makes local state predictions and then algorithmically propagates them through the process. The second, \proglobal, is an end-to-end neural model that makes all state predictions using global information.

%\subsection{Overview}

%We treat process paragraph understanding as correctly tracking participants through it (Section~\ref{sect:end_task}). 
%This requires mechanisms for 
%\begin{ite}
%\item mapping from event sentences to participant states, and 
%\item propagating states from one step to the next (e.g., so states persist if a participant is not mentioned).
%\end{ite}
%These in turn require design decisions for how to encode sentences and represent states in the first place.
%Table~\ref{system-comparison} summarizes these decisions for the two baseline systems we use (QRN and EntNet), and the two 
%models described below.

%We explore two different ways to attack the end task (Section \ref{sect:end_task}). In Section \ref{sect:local_model} we describe \prolocal, a neural model that first makes sentence-level state predictions, the algorithmically propagates them forwards and backwards in time to fill in unknown states. In Section \ref{sect:global_model}, we present \proglobal, an end-to-end neural model that predicts all entity states at each step of the process using the entire paragraph as context.

%\begin{itemize}
%\item What is the overall intuition
%\item How is it encoded in the model
%\item What is the overall architecture
%\item What are individual building blocks
%\item Any specific peculiarities?
%\end{itemize}

\subsection{\prolocal: A Local Prediction Model} 
\label{sect:local_model}

%\bhavana{TODO: Update notations to be in sync with task description and global model.}
%{\textcolor{Red}{PEC: You don't mention how you merge information if there's more than on verb in a sentence} Fixed
%Overall intuition

%\noindent
\eat{
The intuition behind ProLocal is as follows:
\begin{des}
\item[{\bf 1. Local Prediction:}] Infer all the direct effects of individual sentences (e.g., ``roots absorb water'' implies water is now at the roots).
\item[{\bf 2. Commonsense Persistence:}] Algorithmically fill in any remaining unknown states by propagating known values forwards and backwards in time. This second step is equivalent to applying a law of inertia. %: in the absence of information to the contrary, things typically do not change.
\end{des}
}

The design of \prolocal{} consists of two main components: \emph{local prediction} and \emph{commonsense persistence}.
The former infers all direct effects of individual sentences and the latter algorithmically propagates 
known values forwards and backwards to fill in any remaining unknown states.

%\noindent\paragraph{1. Local Prediction}

\subsubsection{Local Prediction}

% Input/Output of the local prediction model
The intuition for local prediction is to treat it as a surface-level QA task. BiLSTMs with span prediction have been effective at answering surface-level questions, e.g., Given ``Roots absorb water.'' and ``Where is the water?'', they can be reliably trained to answer ``Roots'' \cite{Seo2016BidirectionalAF}.  We incorporate a similar mechanism here.

Given a sentence (step) and a participant $e$ in it, the local prediction model makes two types of predictions: 
the change type of $e$ (one of: \textit{no change}, \textit{created}, \textit{destroyed}, \textit{moved}) and the
locations of $e$ before and after this step.  The change type prediction is a multi-class classification problem, 
while the location prediction is viewed as a SQuAD-style surface-level QA task with the goal to find a location span 
in the input sentence.
The design of this model is depicted in Figure~\ref{model1}(a), which adapts a 
bidirectional LSTM~\cite{hochreiter1997long} recurrent neural network architecture (biLSTM)
with attention for input encoding. The prediction tasks are handled by
two different output layers.
We give the detail of these layers below.

\eat{

The input/output behavior is as follows: \bfit{Given} a $step_{i}$ (sentence), participant $e_j$ (span of $step_{i}$), and (via a POS tagger) all verbs $v_{i}$ in that step; \bfit{Predict} the type of change $e_{j}$ undergoes (one of: no change, created, destroyed, moved) {\it and} the before and after locations of $e_{j}$. The change type prediction is implemented as a softmax classification task, and the location prediction is implemented as a span prediction task using BIO tagging.
}

\eat{
\myparagraph{Input Encoding} In Figure \ref{participant-grid} showing step 1, participant ``water'', and verb ``absorb'', each input 
word $w_{ik}$ in $step_i$ is encoded with a vector $x_{ik}=[v_w:v_e:v_v]$, where $[:]$ denotes concatenation, $v_w$ is the pre-trained (GloVe) word embedding,
% \footnote{We use 100 dimensional GloVe embeddings pretrained on Wikipedia 2014 and Gigaword 5 corpora: \url{https://nlp.stanford.edu/projects/glove}}, 
$v_e$ is an indicator variable for the participant ($v_{e}=1$ if $w_{k}$ is ``water'', 0 otherwise), and $v_{v}$ is the an indicator for the verb ($v_{v}=1$ if $w_{k}$ is ``absorb'', 0 otherwise).
}

\myparagraph{Input Encoding} Each word $w_{i}$ in the input sentence is encoded with a vector $x_{i}=[v_w:v_e:v_v]$, the concatenation of 
a pre-trained GloVe \cite{pennington2014glove} word embedding $v_w$, indicator variables $v_e$ on whether $w_{i}$ is the specified participant and $v_v$ on
whether $w_{i}$ is a verb (via a POS tagger). 

\eat{
In Figure \ref{participant-grid} showing step 1, participant ``water'', and verb ``absorb'', each input 
word $w_{ik}$ in $step_i$ is encoded by a vector $x_{ik}=[v_w:v_e:v_v]$, where $[:]$ denotes concatenation, $v_w$ is the pre-trained (GloVe) word embedding,
% \footnote{We use 100 dimensional GloVe embeddings pretrained on Wikipedia 2014 and Gigaword 5 corpora: \url{https://nlp.stanford.edu/projects/glove}}, 
$v_e$ is an indicator variable for the participant ($v_{e}=1$ if $w_{k}$ is ``water'', 0 otherwise), and $v_{v}$ is the an indicator for the verb ($v_{v}=1$ if $w_{k}$ is ``absorb'', 0 otherwise).
}

\myparagraph{Context Encoding} A biLSTM is used to contextualize the word representations in a given sentence. $h_i$ denotes the concatenated output of the bidirectional LSTM for the embedded word $x_{i}$, and encodes the word's meaning in context. % (its contextual embedding).

\myparagraph{Bilinear Attention} Given the participant and verb, the role of this layer is to identify which contextual word embeddings to attend to for generating the output. We first create $h_{ev}$ by concatenating the contextual embedding of the participant and verb.\footnote{Multi-word entities/verbs or multiple verbs are represented by the average word vectors.} %  as a vector using BoW encoder.} 
We then use a bilinear similarity function $sim(h_{i}, h_{ev}) = (h_{i}^T * B * h_{ev}) + b$, 
similar to~\cite{Chen2016ATE}, to compute attention weights $A_{i}$ over each word $w_i$ in the sentence.

\eat{
%As there are two outputs (change type, locations), we train two sets of attention parameters, one for each output. 
We use a bilinear similarity function similar to~\cite{Chen2016ATE} to compute attention weights $A_{i}$ over the contextual embedding $\tilde{x}_{i}$ for each word in the sentence, using the target participant and verb encoded as an entity-verb vector $\tilde{x_{ev}}$ (= their concatenated contextual embeddings)\footnote{Multi-word entities/verbs or multiple verbs are represented as a vector using BoW encoder.}. %Since the dimensions of the contextual embedding matrix and the entity-verb vector are different, 
%Let $\tilde{x_{i}}$ be the contextual embedding matrix for the given $step_i$, and let $\tilde{x_{ev}}$ be the entity-verb vector. The bilinear similarity is then computed using following formula: 
$sim(\tilde{x_{i}}, \tilde{x_{ev}}) = \tilde{x_{i}}^T B \tilde{x_{ev}} + b$.
%During the training phase, the weight matrix $B$ and scalar $b$ are tuned so that the attention layer outputs weights that focus on important parts of the sentence. 
}

%We train two sets of attention parameters one for classification (${B_1, b_1}$) and another for the tagging output (${B_2, b_2}$). 
For state change type prediction, the words between the verb and participant may be important, while for the location tagging, contextual cues such as ``from'' and ``to'' could be more predictive. Hence, we train two sets of attention parameters resulting in weights $A_1$ and $A_2$ which are combined with the contextual vectors $h_{i}$ as described below to produce hidden states $o_1$ and $o_2$ that are fed to the output layers. Here, $|step|$ refers to number of words in the given step or sentence.
\begin{gather*}
o_1 = \sum_i A_{1i} * h_i\\
o_2 = [ (A_{21} * h_1) : (A_{22} * h_2) : \ldots : (A_{2|step|} * h_{|step|})]
\end{gather*}
%{\textcolor{Red}{PEC: I don't understand the below!} \\
%\begin{gather*}
%A1_{ik} = softmax_i (\tilde{x_{i}}^T B_1 \tilde{x_{ev}} + b_1)  \\
%o_1 = \sum_k A1_{ik} * \tilde{x_{ik}}\\
%A2_{ik} = softmax_i ( \tilde{x_{i}}^T B_2 \tilde{x_{ev}} + b_2)  \\
%o_2 = [ (A2_{i1} * \tilde{x_{i1}}) : (A2_{i2} * \tilde{x_{i2}}) : \ldots : (A2_{ic} * %\tilde{x_{ic}})]
%\end{gather*}
\noindent
\myparagraph{Output 1: State Change Type}
We apply a feed-forward network on hidden state $o_1$ to derive the probabilities of the four state change type categories: Create, Destroy, Move and None. % (no change).
%\begin{displaymath}
%State\ change\ type = Softmax(W_t * o_1)
%\end{displaymath}
\eat{
If the predicted change is ``Move'', the before and after locations are then identified from tagging (below). If the change is ``Create'', 
then ``does not exist'' is assigned to the before location
the before location is necessarily ``-'' (does not exist) and only the after location from tagging is used (below). Similarly, for ``Destroy'', the after location is necessarily ``-'', and for ``None'' the after location (if predicted) is considered as both before and after location for the participant.
}

\myparagraph{Output 2: Location Spans} The second output is computed by predicting BIO tags (one of five tags: B-Before-LOC, I-Before-LOC, B-After-LOC, I-After-LOC, O) for each word in the sentence. We apply a feed-forward network on hidden state $o_{2i}$ for $word_i$  to derive the probabilities of these location tags.
%: $tag_i = Softmax(W_l * o_{2i})$\\
Notice that if the change type is predicted as ``Create" (or ``Destroy'') then only the ``after" (or ``before'') location prediction is used.

\myparagraph{Training}
We train the state change type prediction and location tag prediction models jointly, where the loss is 
the sum of their negative log likelihood losses. We use Adadelta~\cite{zeiler12} with learning rate 0.2 to minimize the total loss.

\eat{
We train the state change type prediction and location tag prediction models using negative log likelihood loss. For example, the loss for datapoint with participant $e_j$ in $step_i$ (training example) with gold state change type label $Type_{gold}$ and gold location tags $Tag_{gold_1} \ldots Tag_{gold_{|step_i|}}$, where  $|step_i|$ is the step length is:
\begin{align*}
L_{ij} %&= -\log(P(gold\ labels))\\
&= -log (P(Type_{gold}) * P(Tag_{gold})) \\
%&= -log P(Type_{gold}) - log P(Tag_{gold}) \\
&= -log P(Type_{gold}) - \sum_{\substack{k=1,\\ Tag_{gold_k}\neq `O'}}^{|step_i|} log P(Tag_{gold_k})
\end{align*}
}
%Note that for the ProLocal model only those datapoints are relevant for which participant $e_j$ is a span within $step_i$.
%We use Adadelta with learning rate 0.2 to minimize the total loss: $\sum_{(ij), e_j \in setep_i} L_{ij}$.

\subsubsection{Commonsense Persistence} \label{sect:prolocal-persistence}

The local prediction model will partially fill in the state change grid, showing the direct locational effects of actions (including ``not exists'' and ``unknown location''). To complete the grid, we then algorithmically apply a commonsense rule of persistence that propagates locations forwards and backwards in time where locations are otherwise missing. Figure~\ref{model1}(b) shows an example when applying this rule, where `?' indicates ``unknown location".  This corresponds to a rule of inertia: things are by default unchanged unless told otherwise. If there is a clash, then the location is predicted as unknown.

\begin{figure*}[hbt]
\centering
\includegraphics[width=0.97\textwidth]{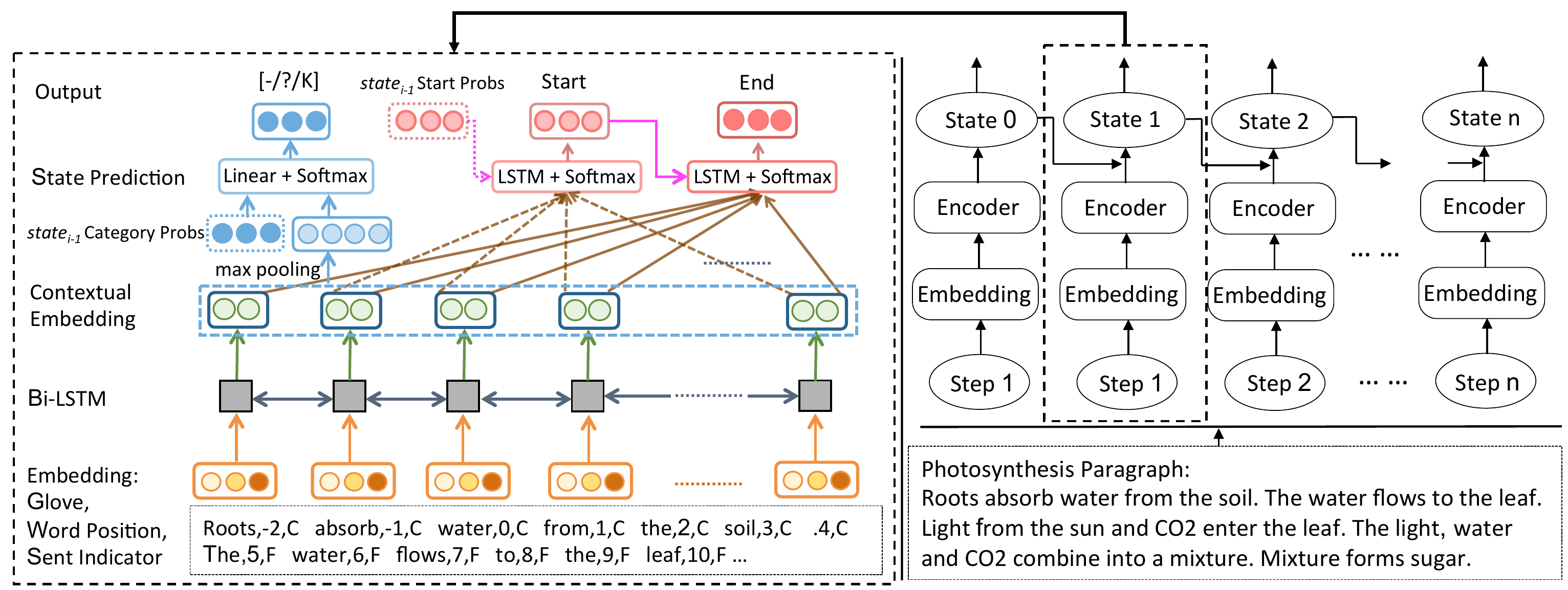}
\caption{\proglobal~predicts a participant's state (type and location) after a given step using bilinear attention over the entire paragraph, combined with its predictions from the previous step.}
\label{model2}
\end{figure*}

\subsection{\proglobal: A Global Prediction Model} \label{sect:global_model}

Unlike \prolocal, the design principle behind \proglobal{} is to model the persistence of state 
information {\it within} the neural model itself, rather than as a post-processing step. % This may improve the robustness of prediction. Thus, 
\proglobal~infers the states of {\it all} participants at each step, even if they are 
not mentioned in the current sentence, using: (1) the global context (i.e., previous sentences), and (2)
the participant's state from the previous step.

\eat{
Unlike \prolocal, the intuition behind \proglobal{} is to model the persistence of state information {\it within} the neural model itself, rather than as a post-processing step. This may improve the robustness of prediction. Thus, \proglobal~infers
%For more than 68\% of the states, the participants or location states are not mentioned in the current sentence, thus  we designed \proglobal{} to jointly
the states of {\it all} participants at each step, even if they are 
not mentioned in the current sentence, using
\vspace{-2mm}
\begin{ite}
\item the global context (i.e., previous sentences),
\item the participant's state from the previous step
\end{ite}
}

%% Input/output

Given a sentence (step) with its context (paragraph) and a participant $e$, \proglobal{}
predicts the existence and location of $e$ after this step in two stages.
It first determines the state of $e$ as one of the classes (``not exist'', ``unknown location'', ``known location'').
%cast as a multi-class classification problem. 
A follow-up location span prediction is made if
the state is classified as ``known location''.

\eat{
To do this, it outputs two items: A state category (not exist ``-'', unknown location ``?'', known location ``K''), implemented as a softmax classification, and a location value (if the category is ``K''), implemented as span prediction.

The input/output behavior of \proglobal~is as follows: \bfit{Given} a $step_{i}$ (sentence) and its context (paragraph $P$), and a participant $e_{j}$,
\bfit{Predict} the existence and location of $e_{j}$ after $step_{i}$. To do this, it outputs two items: A state category (not exist ``-'', unknown location ``?'', known location ``K''),
implemented as a softmax classification, and a location value (if the category is ``K''), implemented as span prediction.
}

Figure~\ref{model2} shows \proglobal's neural architecture, where the left side
is the part for state prediction at each step, and the right side depicts
the propagation of hidden states from one step to the next. %$step_{i-1}$ to $step_i$.
We discuss the detail of this model below.

\eat{
The input/output behavior of \proglobal~is as follows: \bfit{Given} a $step_{i}$ (sentence) and its context (paragraph $P$), and a participant $e_{j}$,
\bfit{Predict} the existence and location of $e_{j}$ after $step_{i}$. To do this, it outputs two items: A state category (not exist ``-'', unknown location ``?'', known location ``K''),
implemented as a softmax classification, and a location value (if the category is ``K''), implemented as span prediction.

\proglobal's neural architecture is shown in Figure~\ref{model2}. Figure~\ref{model2} (left) describes the part of this model for state prediction at each step, 
and the propagation of hidden states from $step_{i-1}$ to $step_i$ is described on the right side of Figure \ref{model2}.
}

\myparagraph{Input Encoding} %To make a state prediction for a participant $e_{j}$ after $step_i$, we feed the entire paragraph into a BiLSTM, with $e_j$ and $step_i$ marked, so the whole paragraph context can be used. To do this marking, 
Given a participant $e$, for each step$_i$, we take the entire paragraph as input. Each word $w$ in the paragraph is represented with three types of embeddings: the general word embedding % $v_w$\footnote{we use the same pre-trained word embeddings as \prolocal}, 
$v_w$,
a position embedding $v_d$ which indicates the relative distance to the participant in the paragraph, and a sentence indicator embedding $v_s$ which shows the relative position (\textit{previous, current, following}) of each sentence in terms of the current step $i$. Both the position embedding and the sentence indicator embedding are of size 50 and are randomly initialized and automatically trained by the model. We concatenate these three types of embeddings to represent each word $x=[v_w:v_d:v_s]$.

\myparagraph{Context Encoding} Similar to \prolocal, we use a biLSTM to encode the whole paragraph % and use $\tilde{h}$ to denote the concatenated output of the bi-LSTM for each word.
and use $\tilde{h}$ to denote the biLSTM output for each word.

\eat{
\myparagraph{State Prediction} The participant's location state can be divided into three categories: ``-'' (does not exist), ``?'' (unknown location) and span (known location). %In order to separate the location span prediction from ``-'' and ``?'', 
Thus, we adopt a two-step prediction strategy: we first classify the location state into one of the three categories (-/?/K), then find the location span if the predicted category is K (known):
% (known location), as follows:
}

\myparagraph{State Prediction} As discussed earlier, we first predict the location state of a participant $e$. 
Let $\tilde{H}_{i}^{P}=[\tilde{h}_{i}^{1}, \tilde{h}_{i}^{2}, ..., \tilde{h}_{i}^{|P|}]$ denote the hidden vectors 
(contextual embeddings) for words in step$_i$ with respect to participant $e$, where $h_{i}^{t}$ denotes the $t$-th word representation output by the biLSTM layer and $P$ is the whole paragraph.  We then apply max pooling to derive a paragraph representation: $\mu_{i}^{P}=\max(\tilde{H}_{i}^{P})$.
To incorporate the category prediction of the previous step, step$_{i-1}$, we concatenate its probability vector $c_{i-1}^{P}\in\mathbb{R}^{3}$
with $\mu_{i}^{P}$, and apply a feed-forward network to derive the probabilities of the three categories:\\ \text{\ \ \ \ } $c_{i}^{P}=\textrm{softmax}(W_c\cdot[\mu_{i}^{P}:c_{i-1}^{P}]+b_c)$

\eat{
\myparagraph{1. Three-Category Classification} Let $\tilde{H}_{ij}^{P}=[\tilde{h}_{ij}^{1}, \tilde{h}_{ij}^{2}, ..., \tilde{h}_{ij}^{|P|}]$ denote the hidden vectors (contextual embeddings) for words in $step_i$ in terms of participant $e_j$, where $h_{ij}^{t}$ denotes the $t^{th}$ word representation output by the bi-LSTM. 
From this, we apply a max pooling function to obtain an overall vector representation for the whole paragraph $\mu_{ij}^{P}=\max(\tilde{H}_{ij}^{P})$. To incorporate the category prediction from $step_{i-1}$, we concatenate the probabilities $c_{(i-1)j}^{P}\in\mathbb{R}^{3}$ from $step_{i-1}$ with $\mu_{ij}^{P}$, and feed the concatenated vector to a linear transformation function followed with a softmax activation to obtain the probabilities for each category:
\[
c_{ij}^{P}=Softmax(W_c\cdot[\mu_{ij}^{P}:c_{i-1j}^{P}]+b_c)
\]
%\noindent
%Thus we can predict whether $e_{j}$ in $step_{i}$ doesn't exist (``-''), exists at an unknown location (``?''), or exists at a known location (``K'').
}

\begin{figure}[bt]
\centering
\includegraphics[width=0.42\textwidth]{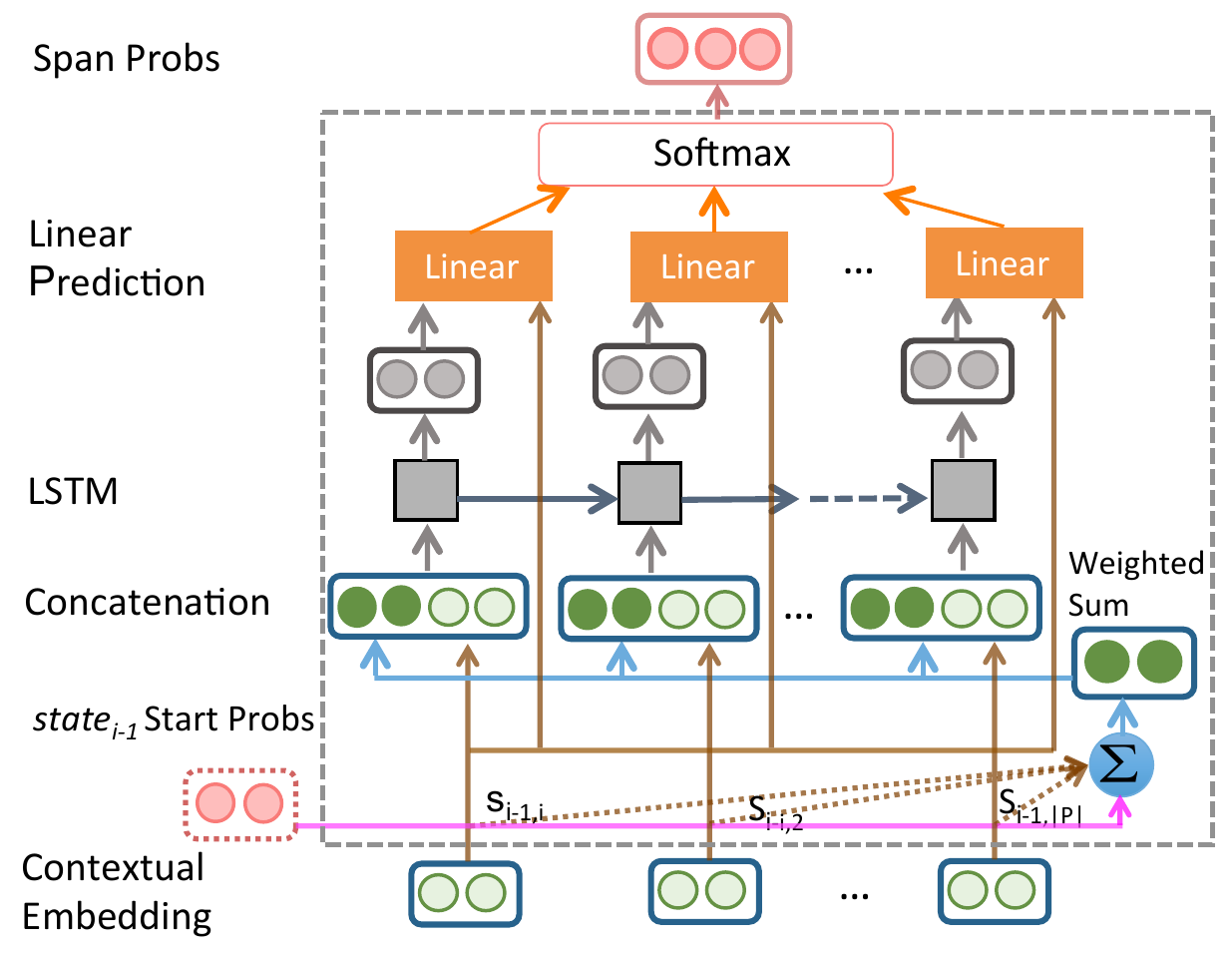}
\caption{Details of the LSTM+Softmax unit, used for predicting the start/end words of a location.}
\label{model2-predict}
\vspace{-0.5cm}
\end{figure}

\myparagraph{Location Span Prediction} (Figure~\ref{model2-predict}). To predict the location span, we predict the start word of the span (by generating a probability distribution over words) and the end word. To predict the location start, we take two types of information as input: the start probability distribution $s_{i-1}^{P} \in\mathbb{R}^{|P|}$ predicted from step$_{i-1}$, and the contextual embeddings $\tilde{H}_{i}^{P}$ of words in the current step$_i$:
\begin{align*}
& \tilde{H}_{i}^{*}= \sum_{t=1}^{|P|}s_{i-1}^{t}\cdot \tilde{H}_{i}^{t}
\\
& \varphi_{i}^{t} = \textrm{LSTM}([\tilde{H}_{i}^{t}:\tilde{H}_{i}^{*}]) 
\end{align*}
where $\tilde{H}_{i}^{*}$ is a sum of word vectors in the paragraph, weighted by the start probabilities from the previous step step$_{i-1}$. $\varphi_{i}^{t}$ is the encoded vector representation for the $t$-th word in the paragraph. %the concatenation of $\tilde{H}_{ij}^{j}$ and $\tilde{H}_{ij}^{*}$.
We then concatenate $\tilde{H}_{i}^{P}$ and $\varphi_{i}^{P}$, and apply a feed-forward network to obtain the start probability distribution for step$_{i}$: $s_{i}^{P}=\textrm{softmax}(W_s\cdot[\tilde{H}_{i}^{P}:\varphi_{i}^{P}] + b_s)$.
Similarly, to predict the end word of the span, we use the start probability distribution $s_{i}^{P}$ of $step_i$ and $\tilde{H}_{i}^{P}$, and apply another LSTM and feed-forward networks to obtain the probabilities. For state$_0$ (the initial location before any steps), we directly feed the sequence of the vectors from the encoding layer to a linear transformation to predict the location start, and apply the same architecture to predict the location end. 
% The linear functions for before and after location predictions are not shared.

%[TODO: add description for the initial bef loc start span prediction]

\begin{table*}
\small
\centering
 \begin{tabular}{|c|c|c|c|c|} \hline
  & Sentence Encoding & Intermediate State Representn. & Propagation through time & Answer Decoding \\
 \hline
EntNet & positional encoding & Dynamic memory blocks & Gated propagation & Softmax classification \\ 
QRN & positional encoding  & Single latent vector $h$ & Gated propagation of $h$ & Softmax classification \\ 
\textsc{ProLocal} & LSTM & Explicit symbolic & Algorithmic & Span prediction \\ 
\textsc{ProGlobal} & LSTM & Distribution over spans & LSTM & Span prediction \\ \hline
 \end{tabular}
\caption{Design decisions in the four neural models.}
\label{system-comparison}
\vspace{-5mm}
\end{table*}

\myparagraph{Training} For each participant $e$ of paragraph $P$, the objective is to optimize the sum of the negative log likelihood of the category classification and location span prediction\footnote{We compute the loss for location span prediction only when the category is annotated as ``known location''.}. We use Adadelta to optimize the models with learning rate 0.5.

%we optimize the category classification and location span prediction using negative log likelihood:
% \begin{displaymath}
% \mathnormal{L}_{j}^{c} = -\frac{1}{N^{'}}\sum_{i=1}^{N^{'}}\log(y_{ij}^{t},c_{ij}^{t})
% \end{displaymath}
% \begin{displaymath}
% \mathnormal{L}_{j}^{loc} = -\frac{1}{2N^{'}}\sum_{t=1, y_{ij}^{t}=known}^{N^{'}}(\log(\hat{y}_{ij}^{t},s_{ij}^{t}) + \log(\tilde{y}_{ij}^{t},\varepsilon_{ij}^{t}))
% \end{displaymath}
% where $N^{'}=N+1$ and $N$ is the total number of steps where location states are known in paragraph $P$. $y_{ij}^{t}$, $\hat{y}_{ij}^{t}$, $\tilde{y}_{ij}^{t}$ are the gold labels for the location category, start and end separately. The overall objective is to optimize the sum of $L^{c}$ and $L^{loc}$. We use Adadelta to optimize the models with learning rate 0.5.

\section{Experiments and Analysis}

%% SY: Please don't kill this line.  I know it's just a filler, but I hate to see a section starts directly with a subsection.  :)
%%     We can comment this out when we don't have other means to save space...
% In this section, we present the experimental results of our models, compared to several baselines.

%%---------------------------------------------------------------------------------------------------------------

% 1. Tasks & Evaluation method

\subsection{Tasks \& Evaluation Metrics}
\label{sec:tasks}

As described in Section~\ref{dataset}, the quality of a model is evaluated based on its ability to
answer three categories of questions, with respect to a given participant $e$:

\noindent
(Cat-1) {\bf Is} $e$ created (destroyed, moved) in the process?

\noindent
(Cat-2) {\bf When} (step\#) is $e$ created (destroyed, moved)?

\noindent
(Cat-3) {\bf Where} is $e$ created (destroyed, moved from/to)?

\eat{
\begin{table}[tbh]
\centering
\boldmath
%\scalebox{0.83}{%
 \begin{tabular}{l|p{5cm}|} 
Category & Template (\# questions) \\
 \hline
Cat-1  & \textbf{Is}  $e_{i}$  created/destroyed/moved?  (750)  \\
Cat-2  & \textbf{When} is  $e_{i}$ created/destroyed/moved? (601) \\
Cat-3  & \textbf{Where} is  $e_{i}$ created/destroyed/moved to/from? (823) \\
\end{tabular}
%}
\caption{Question statistics in test partition of the dataset.}
\label{table:test-statistics}
%\vspace{-0.15in}
\end{table}
}

\noindent
These questions are answered by simple scans over the state predictions for the whole process.
(Cat-1) is asked over all participants, while (Cat-2) and (Cat-3) are asked over just those participants that were created (destroyed, moved).
The accuracy of the answers is used as the evaluation metric, except for questions that may have multiple answers (e.g., ``\emph{When is $e$ moved?}").
In this case, we compare the predicted and gold answers and use the F$_1$ score as the ``accuracy" of the answer set prediction.\footnote{
This approach has been adopted previously for questions with multiple answers (e.g.,~\cite{berant2013webq}).  For questions with only one answer,
F$_1$ is equivalent to accuracy.}

For questions in category (3), an answer is considered correct if the predicted location is identical to, or a sub-phrase of, the labeled location (typically just one or two words), after stop-word removal and lemmatizing.

\subsection{Baseline Methods}

% We also include a baseline for (1) which simply predicts ``not created'' (not destroyed,not moved). 

We compare our models with two top methods inspired by the bAbI dataset, Recurrent Entity Networks (EntNet) and 
Query Reduction Networks (QRN), described earlier in Section~\ref{related-work}. 
Both models make different use of gated hidden states to propagate state information through time, and generate 
answers using softmax. The detailed comparisons in their design are shown in Table~\ref{system-comparison}.

\finaledit{We use the released implementations\footnote{\finaledit{\url{https://github.com/siddk/entity-network} and \url{https://github.com/uwnlp/qrn}}} (with default hyper-parameter values), and retrained them on our dataset, adapted to the standard bAbI QA format.}  
Specifically, we create three separate variations of data by adding three bAbI-style questions after each
step in a paragraph, respectively: 
\begin{enu}
  \item[\hspace{.2in}Q1.] ``Does $e$ exist?'' (yes/no)
  \item[\hspace{.2in}Q2.] ``Is the location of $e$ known?''  (yes/no)
  \item[\hspace{.2in}Q3.] ``Where is $e$?''  (span)
\end{enu}
The template Q1 is applied to all participants. Q2 will only be present in the training data if Q1 is ``yes'',
and similarly Q3 is only present if Q2 is ``yes''. These three variations of data are used to train three
different models from the same method.

\eat{
We used the released implementations and the standard bAbI QA format, applied to our task by authoring bAbI-style questions using three templates:
\begin{des}
\item ``Does $e_{i}$ exist?'' (answer: yes/no), posed after each $step_{j}$ where $j=1...J$ (the number of sentences in paragraph $P$)
\item  ``Is the location of $e_{i}$ known?'' (yes/no), posed after each $step_{j}$.
\item ``Where is $e_{i}$?'' (span), posed after each $step_{j}$.
\end{des}
By definition, Q2 will only be present in the training data if Q1 is ``yes'',
% (it is meaningless to ask about an entity's location if the entity doesn't exist), 
and similarly Q3 is only present if Q2 is ``yes''. 
}

%% Testing
At test time, we cascade the questions (e.g., ask Q2 only if the answer to the Q1 model is ``yes''), 
and combine the model outputs accordingly to answer the questions in our target tasks (Section~\ref{sec:tasks}).

\finaledit{We also compare against a rule-based baseline and a feature-based baseline.
The rule-based method, called ProComp, uses a set of rules that map (a SRL analysis of) each sentence to
its effects on the world state, e.g., IF X moves to Y THEN after: at(X,Y).
The rules were extracted from VerbNet \cite{schuler2005verbnet} and expanded.
% For a full description of ProComp, see 
A full description of ProComp is available at \cite{propara-arxiv}.
The feature-based method uses a Logistic Regression (LR) classifier to predict the
state change type (Move, Create, etc.) for each participant + sentence pair, then a NER-style CRF model
to predict the from/to locations as spans of the sentence. The LR model uses
bag-of-word features from the sentence, along with a discrete feature indicating
whether the participant occurs before or after the verb in the given sentence.
The CRF model uses standard NER features including capitalization, a verb indicator, the previous 3 words,
and the POS-tag of the current and previous word. Similar to our \prolocal{} model, we
apply commonsense persistence rules (Section \ref{sect:prolocal-persistence}) to complete the partial state-change grids predicted by both these baselines.}

%allowing the system to determine the locational state (``-'', ``?'', or a span) of all entities $e_{i}$ and steps $step_{j}$. 
%From this, the end-task below can be performed. Three separate models were trained, one for each question template. 

%%---------------------------------------------------------------------------------------------------------------

% 3. Results

% **************** ORIGINAL VERSION **************

{
\begin{table*}[t]
\centering
\boldmath
\scalebox{0.85}{%
 \begin{tabular}{|l|lcccc|cc|c|} 
\hline
Question type & \multicolumn{5}{c|}{Baseline Models} & \multicolumn{2}{c|} {Our Models} & Human\\
(\# questions) &   Majority  & QRN & EntNet & Rule-based & Feature-based & \prolocal & \proglobal & Upper Bound \\
 \hline
Cat-1 (750) &  51.01 & 52.37 & 51.62 & 57.14 & 58.64 & 62.65 & 62.95 & 91.67 \\
Cat-2 (601) &  - & 15.51 & 18.83 & 20.33 & 20.82 & 30.50 & 36.39 & 87.66 \\
Cat-3 (823) &  - & 10.92 & 7.77 & 2.4 & 9.66 & 10.35 & 35.9 & 62.96 \\
\hline
macro-avg &  - & 26.26 & 26.07 & 26.62 & 29.7 & 34.50 & 45.08 & 80.76 \\
micro-avg &  & 26.49 & 25.96 & 26.24 & 29.64 & 33.96 & 45.37 & 79.69 \\
\hline
 \end{tabular}
}
\caption{\finaledit{Model accuracy on the end task (test partition of \emph{ProPara}). Questions are (Section~\ref{sec:tasks}): (Cat-1) {\bf Is} $e_{i}$ created (destroyed, moved)? (Cat-2) {\bf When} is $e_{i}$ created (...)? (Cat-3) {\bf Where} is $e_{i}$ created (...)?
} }
\label{table:Accuracy-qa-task}
\vspace{-0.15in}
\end{table*}
}

\subsection{Results}
\textbf{Parameter settings:} 
Both our models use GloVe embeddings of size 100 pretrained on Wikipedia 2014 and Gigaword 5 corpora\footnote{\url{https://nlp.stanford.edu/projects/glove}}.  The number of hidden dimensions for the biLSTM are set to 50(\prolocal) and 100(\proglobal). Dropout rates \cite{srivastava2014dropout} for the contextual encoding layer are  0.3(\prolocal) and 0.2(\proglobal). \proglobal{} uses word position and sentence indicator embeddings each of size 50, and span prediction LSTMs with a hidden dimension of 10.
The learning rates for Adadelta optimizer were 0.2(\prolocal) and 0.5(\proglobal). 
\finaledit{Our models are trained on the train partition and the parameters tuned on the dev partition.} 

\eat{
\begin{table}[tbh]
\centering
\boldmath
\scalebox{0.83}{%
 \begin{tabular}{l|cc|cc|} 
Question & \multicolumn{2} {c|} {\prolocal} & \multicolumn{2} {c|} {\proglobal} \\ 
 Type & dev & test & dev & test  \\
 \hline
Cat-1 & 68.52 & 62.65 & 72.12 & 62.95 \\
Cat-2 & 34.27 & 30.50 & 42.26 & 36.39 \\
Cat-3 & 13.10 & 10.35 & 41.32 & 36.59 \\
\hline
Average & 38.63 & 34.50 & 51.90 & 45.31 \\
\end{tabular}
}
\caption{\textbf{Comparing the performance of our models on \emph{ProPara} dev and test partitions.}}
\label{table:Dev-vs-Test}
%\vspace{-0.15in}
\end{table}
}

%\finaledit{Our models are trained using train partition and parameters are tuned based on the performance achieved on the dev partition. Table \ref{table:Dev-vs-Test} compares the performance of \prolocal{} and \proglobal{} models on the dev, test partitions.}

Table~\ref{table:Accuracy-qa-task} compares the performance of various models on the \emph{ProPara} test partition. % , averaging the results for questions in each of the three categories. 
For the first category of questions, we also include a simple majority baseline.
We aggregate results over the questions in each category, and report both macro and micro averaged accuracy scores.
%its three versions (create, destroy, move), or four versions for the last task (created at, destroyed at, moved from, moved to). 

From Table~\ref{table:Accuracy-qa-task}, we can see that EntNet and QRN perform comparably when applied to \emph{ProPara}.  However, despite being the 
top-performing systems for the bAbI task, when predicting whether a participant entity is created, destroyed or moved, their
predictions are only slightly better than the majority baseline. Compared to our local model \prolocal{}, EntNet and QRN are
worse in predicting the exact step where a participant is created, destroyed or moved, but better in predicting the location.
The weak performance of EntNet and QRN on \emph{ProPara} is understandable:  both systems were designed 
with a different environment in mind, namely a large number of examples from a few conceptual domains~(e.g., 
moving objects around a house), covering a limited vocabulary.  As a result, they might not scale well
when applied to real procedural text, which justifies the importance of having a real challenge dataset like \emph{ProPara}.

\finaledit{
Although the rule-based baseline \cite{propara-arxiv} uses rules mapping SRL patterns to state changes,
its performance appears limited by the incompleteness and approximations in the rulebase, and by errors by the SRL parser.
The feature-based baseline performs slightly better, but its performance is still poor compared to our neural models. This
suggests that it has not generalized as well to unseen vocabulary
(25\% of the test vocabulary is not present in the train/dev partitions of \emph{ProPara}).}

\eat{
\textbf{Rule-based baseline \cite{propara-arxiv} has access to high-quality VerbNet style semantic lexicon that maps syntactic patterns around verbs to state changes. However, its performance is severely limited by lexicon coverage and the ability of SRL parser to map  an input sentence into a syntactic pattern in the lexicon. Feature-based method makes independent decisions about the state change type and location values. Even though it outperforms rule-based method its performance is poor when compared to our proposed neural models. This could be due the fact that 25\% of the test vocabulary was unseen in the train/dev partitions of \emph{ProPara}. It is known that neural models tend to generalize well to unseen data.}
}

When comparing our two models, it is interesting that \proglobal{} performs substantially better than \prolocal. 
One possible cause of this is cascading errors in \prolocal: if a local state prediction is wrong, it may still be propagated 
to later time steps without any potential for correction, thus amplifying the error. In contrast, \proglobal~makes a state 
decision for every participant entity at every time-step, taking the global context into account, and thus appears more robust to cascading errors.
\finaledit{Furthermore, \proglobal's gains are mainly in Cat-2 and Cat-3 predictions, which rely more heavily on out-of-sentence cues. For example, 30\% of the time the end-location is not explicitly stated in the state-change sentence, meaning \prolocal~cannot predict the end-location in these cases (as no sentence span contains the end location). \proglobal, however, uses the entire paragraph and may identify a likely end-location from earlier sentences.}
% \finaledit{Further \proglobal's gains are mainly evident in Cat-2 and Cat-3 predictions since they rely more heavily on out-of-sentence cues. E.g., 30\% of the time the end-location is not explicitly stated in the state-change sentence, causing significant problems for \prolocal. Presence of a state-change (Cat-1 questions), though, is more directly related to local information (in particular the verb).}

\finaledit{Finally, we computed a human upper bound for this task (last column of Table \ref{table:Accuracy-qa-task}). During dataset creation, each grid was fully annotated by 5 different Turkers (Section~\ref{dataset}). Here, for each grid, we identify the Turker whose annotations result in the best score for the end task with respect to the other Turkers' annotations. The observed upper bound of $\sim$80\% suggests that the task is both feasible and well-defined, and that there is still substantial room for creating better models.}
% . Also, there is a scope of improvement for the existing RC models (proposed \proglobal{} model achieves on average 45\% for this task).  }

%%---------------------------------------------------------------------------------------------------------------

\subsection{Analysis}
\label{subsec:analysis}

\begin{figure}[t]
\centering
\includegraphics[width=1.0\columnwidth]{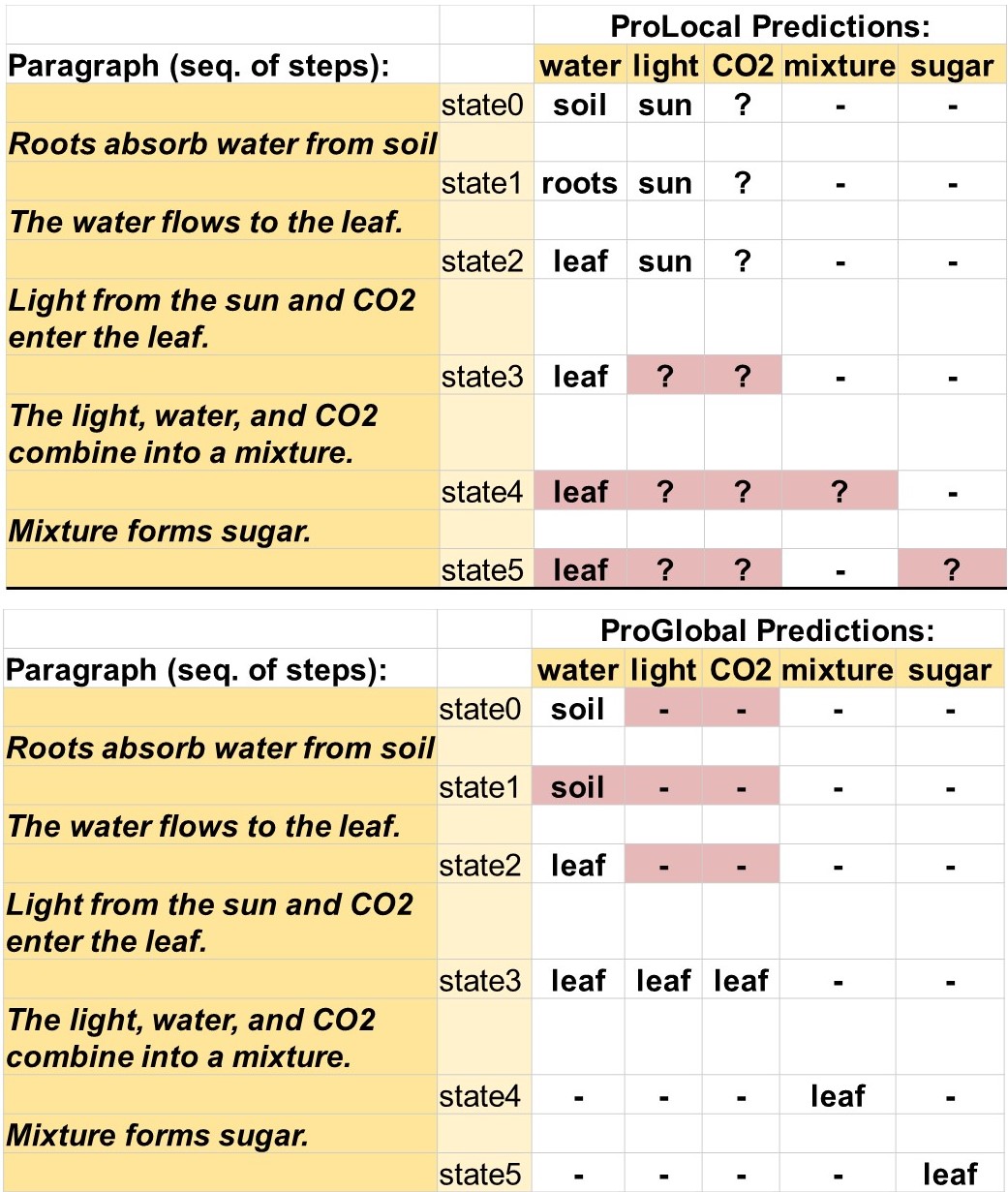}
%\caption{Results for the photosynthesis process example: the cells with red background color show the prediction errors.}
\caption{\prolocal~(top) and \proglobal~(bottom) predictions on a simple paragraph (errors in red).}
\label{exampleResult}
\vspace{-3mm}
\end{figure}

To further understand the strengths and weaknesses of our systems, we ran the simplified paragraph in Figure~\ref{participant-grid} 
verbatim through the models learned by \prolocal{} and \proglobal. The results are shown in Figure~\ref{exampleResult}, 
with errors highlighted in red.

\prolocal{} correctly interprets ``\textit{Light from the sun and CO2 enters the leaf.}'' to imply that the light was at the sun before the event. In addition, as there were no earlier mentions of light, it propagates this location backwards in time, (correctly) concluding the light was initially at the sun. 
% (This backwards propagation in time is not modeled by the other systems). 
However, it fails to predict that ``\textit{combine}'' (after state 3) destroys the inputs, resulting in continued prediction of the existence and locations for those inputs. One contributing factor is that \prolocal's predictions ignore surrounding sentences (context), potentially making it harder to distinguish destructive vs. non-destructive uses of ``\textit{combine}''. 

\proglobal{} also makes some errors on this text, most notably not realizing the light and CO2 exist from the start (rather, they magically appear at the leaf). Adding global consistency constraints may help avoid such errors. It is able to predict the sugar is formed at the leaf, illustating its ability to persist and transfer location information from earlier sentences to draw correct conclusions.
% (here two sentences earlier) to draw correct conclusions.

We additionally 
randomly selected 100 prediction errors from the dev set for \proglobal, and identified four phenomena contributing to errors:

\textbf{(1) Implicit Creation/Destruction:} In 37\% of the errors, the information about the creation or destruction of a participant is implicit or missing, which resulted in existence classification errors. For example, in the sentences ``\textit{A fuel goes into the generator. The generator converts mechanical energy into electrical energy.}'', ``\textit{fuel}'' is implicitly consumed as the generator converts mechanical energy into electrical energy.

\textbf{(2) Location Errors:} In 27\% of the examples, the location spans were not perfectly identified as follows: absolute wrong location span prediction (17\%), longer span prediction (6\%), and location prediction from different granularity (4\%). 

\textbf{(3) Complex Syntax:} In 13\% of the examples, a moving participant and its target location are separated with a wide context within a sentence, making it harder for the model to locate the location span. 

\textbf{(4) Propagation:} \proglobal{} tends to propagate the previous location state to next step, which may override locally detected location changes or propagate the error from previous step to next steps. 9\% of the errors are caused by poor propagation.

\subsection{Future Directions}

This analysis suggests several future directions:
\textbf{Enforcing global consistency constraints:} e.g., it does not make sense to create an already-existing entity, or destroy a non-existent entity.
  Global constraints were found useful in the earlier ProRead system \cite{berant2014modeling}.\\
\textbf{Data augmentation through weak supervision: }additional training data can be generated
  by applying existing models of state change, e.g., from VerbNet \cite{kipper2008large}, to new sentences
  to create additional sentence+state pairs.\\
%   (\cite{bosselut2017simulating} similarly use rule-based weak supervision to generate state-change training data).
\textbf{Propagating state information backwards in time:} if $e_j$ is at $l_{ij}$ after $step_i$, it is likely to
  also be there at $step_{i-1}$ given no information to the contrary. \proglobal, EntNet, and QRNs are inherently unable to
  learn such a bias, given their forward-propagating architectures.

\vspace{-1mm}
\section{Conclusion}
\vspace{-1mm}
New datasets and models are required to take reading comprehension to a deeper level of machine understanding.
As a step in this direction, we have created the ProPara dataset, the first to contain 
natural text about a changing world along with an annotation of entity states 
during those changes. We have also shown that this dataset presents new challenges for previous 
models, and presented new models that exploit ideas from surface-level QA, in particular
LSTM input encoding and span prediction, producing performance gains.
\finaledit{The dataset and models are available at \url{http://data.allenai.org/propara}.}
\vspace{1mm}

\noindent
\eat{
\finaledit{{\bf Acknowledgements:} We are grateful to Oren Etzioni, Carissa Schoenick, Mark Neumann, and Isaac Cowhey for their critical contributions.}
}
% Future directions include exploiting global consistency constraints, 
% adding more data through distant supervision, and extending \proglobal~to
% propagate state knowledge backwards as well as forwards in time.
% We are releasing ProPara and the models to the community.
% Despite this, there is still substantial room for prediction
% improvement, e.g., adding distant supervision, or enforcing process consistency constraints.
% We look forward to continued progress in this area.

\section*{Acknowledgements}
\vspace{-2mm}
\finaledit{
We are grateful to Paul Allen whose long-term vision continues to inspire our scientific endeavors. We also thank Oren Etzioni, Carissa Schoenick, Mark Neumann, and Isaac Cowhey for their critical contributions to this project.
}

\bibliography{references}
\bibliographystyle{acl_natbib}
\end{document}